# Image enhancement using fusion by wavelet transform and laplacian pyramid

S.M.Mukane[1], Y.S.Ghodake[2] and P.S.Khandagle[3]

[1] Department of Electronics & Telecommunication
SVERI's, College of Engineering, Pandharpur

[2] Department of Electronics & Telecommunication
SVERI's, College of Engineering, Pandharpur

[3] Department of Electronics & Telecommunication
SSPM's College of Engineering, Kankavali

**Abstract**
The idea of combining multiple image modalities to provide a single, enhanced image is well established different fusion methods have been proposed in literature. This paper is based on image fusion using laplacian pyramid and wavelet transform method. Images of same size are used for experimentation. Images used for the experimentation are standard images and averaging filter is used of equal weights in original images to burl. Performance of image fusion technique is measured by mean square error, normalized absolute error and peak signal to noise ratio. From the performance analysis it has been observed that MSE is decreased in case of both the methods where as PSNR increased, NAE decreased in case of laplacian pyramid where as constant for wavelet transform method.

## 1. Introduction
In image fusion is the process of combining relevant information from two or more images into a single image. The resulting image will be more enhanced than any of the input images. The concept of image fusion has been used in wide variety of applications like medicine, remote sensing, machine vision, automatic change detection, bio metrics etc. With the emergence of various image-capturing devices, it is not possible to obtain an image with all the information. Sometimes, a complete picture may not be always feasible since optical lenses of imaging sensor especially with long focal lengths, only have a limited depth of field. Image fusion helps to obtain an image with all the information. Image fusion is a concept of combining multiple images into composite Products, through which more information than that of individual input images can be revealed.

## 2. Motivation:
The concept of data fusion goes back to the 1950's and 1960's with the search for practical methods of merging images from various sensors to provide a composite image which could be better identify natural and manmade objects. In the past decade, medical imaging, night vision, military and civilian avionics, autonomous vehicle navigation, remote sensing, concealed weapons detection and various security and surveillance systems are only some of the applications that have benefited from such multi sensor arrays [3]. Motivation for image fusion is mainly the step toward recent technological advances in the fields of image fusion technique method. Improved robustness and increased resolution of modern imaging sensors and, more significantly, availability at a lower cost, have made the use of multiple sensors common in a range of imaging applications [4]. In this image fusion technique using laplacian pyramid and wavelet transform method increases the image quality of a fused image.GUI of these implemented method [10].

## 3. Image fusion techniques theory
These techniques both method having the two input images and these two images with laplacian pyramid and wavelet transform are used for fusion. From these two methods of image fusion we get a single fused image which has better quality and more enhanced image.

3.1 Image fusion by using dwt
The wavelet transform decomposes the image into low-high, high-low, high-high spatial frequency bands at different scales and the low-low band at the coarsest scale. The L-L band contains the average image information whereas the other bands contain directional information due to spatial orientation. Higher absolute values of wavelet coefficients in the high bands correspond to salient features such as edges or lines [7]. First the images are transformed to the wavelet domain with the function dwt2.m; the dwt2 command performs single-level two-dimensional wavelet decomposition with respect to either a particular ('wname','type') or particular wavelet decomposition filters (Lo-D and Hi-D) show in fig.1

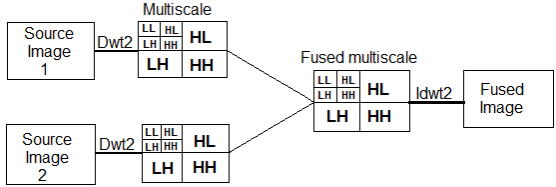

Figure.1 schematic of wavelet based fusion

The DWT is applied to both images and a decomposition of each original image is achieved. This is represented in the multi scale illustration where different bars (horizontal, vertical, diagonal and none) represent also different coefficients. The different boxes, associated to each decomposition level, are coefficients corresponding to the same image spatial representation in each original image, i.e. the same pixel or pixels positions in the original images. Only coefficients of the same level and representation are to be fused, so that the fused multi scale coefficients can be obtained [5]. This is displayed in the diagonal details where the curved arrows indicate that both coefficients are merged to obtain the new fused multi scale coefficients. This is applicable to the remainder coefficients. Once the fused multi scale is obtained, through the IDWT, the final fused image is achieved [6].

3.2 Image fusion by using Laplacian Pyramid
The image pyramid is a data structure designed to support efficient scaled convolution through reduced image representation. It consists of a sequence of copies of an original image in which both sample density and resolution are decreased in regular steps. These reduced resolution levels of the pyramid are themselves obtained through a highly efficient iterative algorithm. The bottom, or zero level of the pyramid, $A_0$, is equal to the original image [2]. This is low pass- filtered and sub sampled by a factor of two to obtain the next pyramid level, $A_1$. $A_1$ is then filtered in the same way and sub sampled to obtain $A_2$. Further repetitions of the filter/subsample steps generate the remaining pyramid levels. To be precise, the levels of the pyramid are obtained iteratively as follows.
For 0 < l < N:
$Al(i, j) \sum \sum w(m, n) Al - 1(2i + m, 2j + n)$
(1)

However, it is convenient to refer to this process as a standard DECREASE operation, and is given by $A_l$ = DECREASE [$A_l$-1]. We call the weighting function w (m, n) in equation 1 as the "generating kernel." For reasons of computational efficiency this should be small and separable. A five-tap filter was used to generate the Gaussian pyramid in Figure.2. Pyramid construction is equivalent to convolving the original image with a set of Gaussian-like weighting functions. The convolution acts as a low pass filter with the band limit reduced correspondingly by one octave with each level. Because of this resemblance to the Gaussian density function we refer to the pyramid of low pass images as the "Gaussian pyramid."

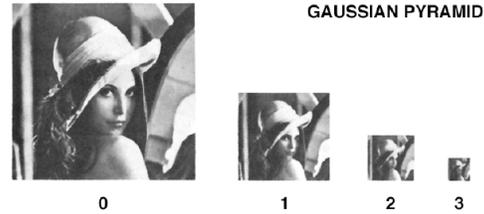

Fig.2 First six levels of the Gaussian pyramid for the "Lady" image. [11]

The original image, level 0, measures 256 by 256 pixels and each higher level array is roughly half the dimensions of its predecessor. Thus, level 5 measures just 8 by 8 pixels [11]. Band pass, rather than low pass, images are required for many purposes. These may be obtained by subtracting each Gaussian (low pass) pyramid level from the next lower level in the pyramid. Because these levels differ in their sample density it is necessary to interpolate new sample values between those in a given level before that level is subtracted from the next-lower level. Interpolation can be achieved by reversing the DECREASE process. We call this an INCREASE operation. Let $A_{l,k}$ be the image obtained by Increasing $A_l$, k times.

Then

$A_{l,k}$ = INCREASE [A $A_{l,k-1}$]

Or, to be precise,

$A_{l,0} = A_l$, and for k>0,

$A(l, k)(i, k) = 4 \sum_m \sum_n A(l, k - 1)(2i + \frac{m}{2}, 2j + \frac{n}{2})$
(2)

Here in equation 2 only terms for which (2i+m)/2 and (2j+n)/2 are integers contribute to the sum. The INCREASE operation doubles the size of the image with each iteration, so that $A_{l,1}$, is the size of $A_{l,1}$ and $A_{l,1}$ is the same size as that of the original image. The levels of the band pass pyramid, $B_0$, $B_1$... $B_n$ may now be specified in terms of the low pass pyramid levels as follows:

$B_l = A_l - \text{INCREASE}[A_{l+1}]$

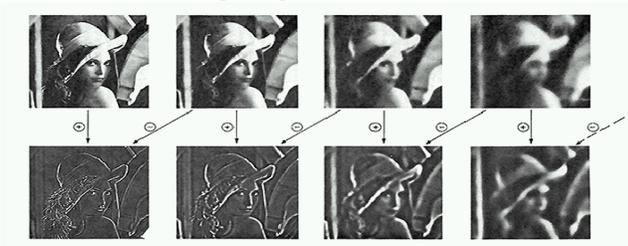

Figure.3 Generation of Laplacian images of 4 levels. [11]

The first four levels are shown in Figure.3 Just as the value of each node in the Gaussian pyramid could have been obtained directly by convolving a Gaussian like equivalent weighting function with the original image, each value of this band pass pyramid could be obtained by convolving a difference of two Gaussians with the original image. These functions closely resemble the laplacian operators commonly used in image processing for this reason we refer to the band pass pyramid as a "Laplacian pyramid." An important property of the laplacian pyramid is that it is a complete image representation: the steps used to construct the pyramid may be reversed to recover the original image exactly. The top pyramid level, $B_n$, is first expanded and added to $B_{n-1}$ to form $A_{n-1}$ then this array is expanded and added to $B_{n-2}$ to recover $A_{n-2}$.

## 4. Experimental result and discussion

This experimental result and discussion depicts implementation for wavelet transforms and implementation of Laplacian pyramid

4.1 Implementation for wavelet transforms.
The discrete wavelets transform (DWT) allows the image decomposition in different kinds of coefficients preserving the image information [5].Such coefficients coming from different images can be appropriately combined to obtain new coefficients, so That the information in the original images is collected appropriately Once the coefficients are merged, the final fused image is achieved through the inverse discrete wavelets transform (IDWT), where the information in the merged coefficients is also preserved [1].The key step in image fusion based on wavelets is that of coefficients combination, namely, the process of merge the coefficients in an appropriate way in order to obtain the best quality in the fused image. This can be achieved by a set of strategies. The most simple is to take the average of the coefficients to be merged [6].filter used in wavelet is db1

4.2 Implementation for laplacian pyramid.
Implemented laplacian pyramid fusion first construct the laplacian pyramid of two input images to be fused. Take the average of the two pyramids corresponding to each level and sum them. The resulting image is simple average of two low resolution images at each level. Decoding of an image is done by expanding, then summing all the levels of the fused pyramid which is obtained by simple averaging. A more efficient procedure is to expand $B_n$ once and add it to $B_n - 1$, then expand this image once and add it to $B_n - 2$, and so on until level 0 is reached and $g_0$ is recovered. This procedure simply reverses the steps in Laplacian pyramid generation. The input arguments of this function are: Source images (im1, im2): must have the same size, and are supposed to be same size example both image of (m*n) matrix size. Number of scales (ns): an integer that defines the number of pyramid decomposition levels. The reduce operation is a two-dimensional convolution with the Gaussian filter followed by a down sampling by two. Expand, as the opposite operation of reduce, performs an up sampling by two followed by a two dimensional convolution with the same Gaussian filter [8]. Steps followed for fusion is given below table 1.

Table 1: steps for Laplacian pyramid fusion

| a1l = reduce (image1) , a2l = reduce (image2) |
| --- |
| a1l-1 = expand (a1l) , a2l-1 = expand (a2l) |
| B1 = image1 – a1l-1 ,  B2 = image2 – a2l-1 |
| B = maximum (B1, B2) |
| Last level of the pyramid, fusion1 = average (a1l, a2l) |
| Reconstruct fusion image, Fusion = expand (fusion1) + B |

4.3 Performance analysis
Table 2 shows data base of input images and fused image. Image1 rice.png and image2 cameraman.tif are input images for fusion. Image3 and image4 fused images by wavelet transform, like wise image 5 and image6 fused images by laplacian pyramid method. Table 3 and 4 shows performance analysis. Performance analysis of the original images and fused image can be identify by three performance metrics Mean Square error, Normalized Absolute Error and Peak Signal To Noise Ratio shows in table 3 and 4

Table 2: Data base of input and fused images

| In put images Image1(rice.png) 256*256 Image2 (Cameraman.tif) 256*256 | 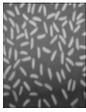 Image1 | 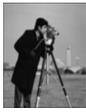 Image2 |
|---|---|---|
| Fused images by wavelet transform. | 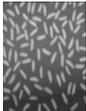 Image3 | 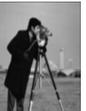 Image4 |
| Fused images using laplacian pyramid. | 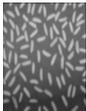 Image5 | 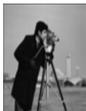 Image6 |

Table 3: performance metrics of cameraman .tif

| Parameters | Results for cameraman.tif image | | |
|---|---|---|---|
| | Image2 | Fused image using wavelet | Fused image using Laplacian |
| MSE | 851.17 | 645.5678 | 232.981 |
| NAE | 0.2501 | 0.2501 | 0.1310 |
| PSNR | 18.869 | 20.0654 | 24.4916 |

Table 4: performance metrics of rice.png

| Parameters | Results for rice.png image | | |
|---|---|---|---|
| | Image1 | Fused image using wavelet | Fused image using Laplacian |
| MSE | 1.0701e+03 | 684.165 | 283.9938 |
| NAE | 0.2498 | 0.2498 | 0.1298 |
| PSNR | 17.8706 | 19.8132 | 23.6317 |

## 5. Conclusion

Method implemented for laplacian pyramid and wavelet transform fusion is gives reliable fused image than the original image for finding the enhancement of the image taken parameter was mean square error, normalized absolute error of fused image reduce both kind of error and increase the peak signal to noise ratio. Image fusion is very much necessary in lot of applications to obtain the high quality image from two source images. There are different image fusions techniques are leading now days but in this seminar we have considered laplacian pyramid. Image fusion using laplacian pyramid and wavelet transform has been simulated using MATLAB software.

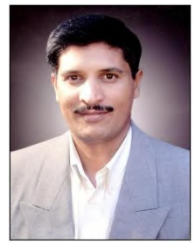

**Shailendra M. Mukane**

Prof. Dr. Shailendra M. Mukane has completed B.E. Electronics Engineering (1993), M.Tech Power Electronics from VTU with Gold Medal (2001) and Ph.D. Electronics Engineering from Solapur University (2013). His total experience is 19 years and research areas include signal processing, image processing, and power electronics. He is fellow member of IEI Kolkata, Life member of ISTE and Associate Member of IEEE.

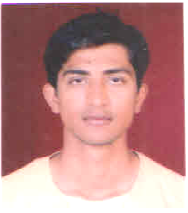

**Yogesh S Ghodake**

Yogesh S Ghodake has completed B.E Electronics and Telecommunication Engineering (2011), pursuing M.E Electronics and Telecommunication in SVERI's College of Engineering Pandharpur.